\documentclass[journal]{IEEEtran}
\usepackage{hyperref}       % hyperlinks
\usepackage{url}            % simple URL typesetting
\usepackage{booktabs}       % professional-quality tables
\usepackage{amsfonts}       % blackboard math symbols
\usepackage{nicefrac}       % compact symbols for 1/2, etc.
\usepackage{microtype}      % microtypography
\usepackage{amsmath}
\usepackage{amsthm}
\usepackage{multirow}
\usepackage{amssymb}
\usepackage{bm} 
\usepackage{graphicx}
\usepackage{subfigure}
\usepackage{colortbl,booktabs}
\usepackage{algorithm}
\usepackage[noend]{algpseudocode}
\usepackage{float}
\usepackage[margin=1.9cm]{geometry}
\usepackage{xcolor}
\definecolor{mygray}{gray}{.9}

\newtheorem{prop}{Proposition}

\begin{document}
\title{Von Mises-Fisher Elliptical Distribution}
\author{Shengxi~Li,~\IEEEmembership{Student~Member,~IEEE},~Danilo~Mandic,~\IEEEmembership{Fellow,~IEEE}
\thanks{Shengxi Li and Danilo Mandic are with the Department of Electrical and Electronic of Imperial College London.}}
\maketitle
\begin{abstract}
A large class of modern probabilistic learning  systems assumes symmetric distributions, however, real-world data tend to obey skewed distributions and are thus not always adequately modelled through symmetric distributions. To address this issue, elliptical distributions are increasingly used to generalise symmetric distributions, and further improvements to skewed elliptical distributions have recently attracted much attention. However, existing approaches are either hard to estimate or have complicated and abstract representations. To this end, we propose to employ the von-Mises-Fisher (vMF) distribution to obtain an explicit and simple probability representation of the skewed elliptical distribution. This is shown not only to allow us to deal with non-symmetric learning systems, but also to provide a physically meaningful way of generalising skewed distributions. For rigour, our extension is proved to share important and desirable properties with its symmetric counterpart. We also demonstrate that the proposed vMF distribution is both easy to generate and stable to estimate, both theoretically and through examples.
\end{abstract}

\section{Introduction}
Probabilistic distributions are a common underpinning tool in modelling, understanding and predicting a wide variety of real-world signals. The normal distribution has been a workhorse in probabilistic modelling, as it admits a simple representation and mathematical tractability, while its application is justified through the central limit theorem. However, issues such as the lack of  robustness and flexibility when dealing with general signals remain a serious obstacle to real-world applications. The family of elliptical distributions generalises normal distributions, and possesses many desired properties such as simple generation, controllable robustness and flexibility. Elliptical distributions include the normal, Cauchy, \textit{t}, logistic, and Weibull distributions \cite{fang2018symmetric}. The well-behaved nature of elliptical distributions underpins powerful modelling tools, such as unimodal \cite{renaux2006high, zozor2018maximum}, mixture models \cite{li2020solving, li2020universal}, Bayesian frameworks \cite{zozor2009map} and probabilistic graphical models \cite{stankovic2020data3}.

Despite success, elliptical distributions inherit some of the limitations of symmetric distributions, which limits their modelling power, as in many cases such as financial, biometric and audio scenarios, the data are not symmetric due to intrinsic coupling, systematic trend, outliers, or a small number of samples available. To address this issue, several skewed versions of the elliptical distributions have been proposed, with the majority \cite{azzalini1999statistical, genton2005generalized, wang2004skew, branco2001general} following a similar way of generalising the normal distribution by adding a skewness weighting function \cite{azzalini1996multivariate}. Although attracting attentions in recent applications \cite{nurminen2015robust, nurminen2018skew, zhang2020extended}, this type of skewed elliptical distributions results in a different stochastic representation from the (symmetric) elliptical distribution, which is prohibitive to invariance analysis, sample generation and parameter estimation. 
A further extension employs a stochastic representation of elliptical distributions by assuming some inner dependency \cite{frahm2004generalized}. However, due to the added dependency, the relationships between parameters become unclear and sometimes interchangeable, impeding the estimation. 

In this paper, we start from the stochastic representation of elliptical distributions, and propose a novel generalisation by employing the von Mises-Fisher (vMF) distribution to explicitly specify the direction and skewness, whilst keeping the independence among the components in elliptical distributions. Such generalisation is intuitive and maximally resembles the original (symmetric) elliptical distributions, which is  beneficial in three aspects: i) it admits a simple and closed-form density function, so that all the elliptical distributions can be explicitly generalised as the proposed vMF elliptical distribution; ii) it shares many desirable properties with the original elliptical distribution, including the independence between the quadratic term (or the Mahalanobis distance) and the whitened variables, the invariance property, and explicit moments; iii) it shares the robustness properties of the elliptical distributions, and can be estimated stably and efficiently, even by a naive numerical gradient descent method.
This opens a new avenue for the design and implementation of robust probabilistic learning systems, such as generative models in unsupervised learning and discriminative models in supervised learning systems.

\section{Existing Generalised Elliptical Distributions}
A random variable $\bm{\mathcal{X}}_e\in\mathbb{R}^m$ is said to satisfy an elliptical distribution when it has the following stochastic representation
\begin{equation}\label{eq_stochastic}
\bm{\mathcal{X}}_e =^d \bm{\mu} + \mathcal{R}\bm{\Lambda}\bm{\mathcal{U}},
\end{equation}
where $\mathcal{R}\in\mathbb{R}$ is a non-negative scalar random variable and $\bm{\mathcal{U}}\in\mathbb{S}^{m'-1}$ is a random variable that is uniformly distributed on a unit sphere surface, i.e., $\mathbb{S}^{m'-1}\!:=\{\mathbf{x}\in\mathbb{R}^{m'}\!:\mathbf{x}^T\mathbf{x}=1\}$. Moreover, $\bm{\mu}\in\mathbb{R}^m$ and $\bm{\Lambda}\in\mathbb{R}^{m\times m'}$ are two constant parameters that control the distribution centres and scatter. It needs to be pointed out that the elliptical distribution is symmetric about its centre, $\bm{\mu}$. This is due to the fact that $\mathcal{R}$ and $\bm{\mathcal{U}}$ in \eqref{eq_stochastic} are independent random variables, thus constituting a spherical distribution around $\mathbf{0}$ via $\mathcal{R}\bm{\mathcal{U}}$. The constant, $\bm{\Lambda}$, transforms the sphere into an ellipse (still centred around $\mathbf{0}$), while $\bm{\mu}$ translates the centre of the elliptical distribution. 

When the cumulative density function (cdf) of $\mathcal{R}$ is absolutely continuous and $\mathbf{\Sigma} = \mathbf{\Lambda}\mathbf{\Lambda}^T$ is non-singular, we can write the probability density function (pdf) of the elliptical distribution as 
\begin{equation}\label{eq_symmpdf}
p_{\bm{\mathcal{X}}_e}(\mathbf{x}) = \mathrm{det}(\mathbf{\Sigma})^{-1/2} \cdot c_m \cdot  g\big((\mathbf{x} - \bm{\mu})^T \mathbf{\Sigma}^{-1} (\mathbf{x} - \bm{\mu})\big),
\end{equation}
where $c_m = \nicefrac{\Gamma(\nicefrac{m}{2})}{2\pi^{\nicefrac{m}{2}}}$ is a constant solely determined by the dimension, $m$, while $g(t)$ ($t=(\mathbf{x} - \bm{\mu})^T \mathbf{\Sigma}^{-1} (\mathbf{x} - \bm{\mu})$) is called the density generator, which is related to the pdf of $\mathcal{R}$ in \eqref{eq_stochastic} \cite{fang2018symmetric}. We denote $\bm{\mathcal{X}}_e$ by $\bm{\mathcal{X}}_e\!\sim\!\mathcal{E}(\bm{\mu},\mathbf{\Sigma},g)$.

The skewness can be achieved by adding a weighting term $\pi(\mathbf{x}-\bm{\mu})$ in \eqref{eq_symmpdf} \cite{azzalini1999statistical, genton2005generalized, wang2004skew}, in a way similar to the skewed normal distribution \cite{azzalini1996multivariate}. However, this type of skewness does not necessarily start from a stochastic representation, which impedes clear interpretations of its inner relationships, generations and moments. A further successful variant employs conditional distributions of a symmetric elliptical distribution \cite{branco2001general}, to give
\begin{equation}\label{eq_skewsto}
\bm{\mathcal{X}}_{se}=^d\bm{\mathcal{Y}}~|~\mathcal{Y}_0 > 0, ~\mathrm{where}~ \begin{bmatrix}
\bm{\mathcal{Y}}\\
\mathcal{Y}_0
\end{bmatrix}\sim\mathcal{E}\big(\begin{bmatrix}
\bm{\mu}\\
0
\end{bmatrix}, \begin{bmatrix}
\mathbf{\Sigma} & \bm{\beta}^T\\
\bm{\beta} & 1
\end{bmatrix},g\big)
\end{equation}
where the parameter $\bm{\beta}$ controls the skewness of the distribution. The form of \eqref{eq_skewsto} represents a typical skewed elliptical distribution, of which $\mathcal{Y}_0$ has also been extended to higher dimensions \cite{sahu2003new,arellano2010multivariate,abtahi2013new}. Importantly, the form of \eqref{eq_skewsto} is invariant under quadratic forms \cite{genton2005generalized}, and is closed under marginalisation and affine transforms; we refer to \cite{adcock2020selective} for more detail. However, the estimation of the above skewed elliptical distributions can be ill-posed, especially regarding its shape (skewness) parameter. Although the singularity issue in the information matrix of shape parameter can be relieved by a centralised parametrisation trick \cite{azzalini1999statistical}, the estimation of the shape parameter in this skewed version could still diverge, which calls for other penalty techniques \cite{azzalini2013maximum}. Note that the moment estimation method employed to estimate skewed normal distributions is also inadequate for skewed elliptical distributions \cite{flecher2009estimating}. 

A further generalisation that explicitly comes with the stochastic representations was proposed by Frahm \cite{frahm2004generalized}
\begin{equation}\label{eq_genlpdf}
\bm{\mathcal{X}}_{ge} =^d \bm{\mu} + \hat{\mathcal{R}}\bm{\Lambda}{\bm{\mathcal{U}}},
\end{equation}
and has a form similar to that in \eqref{eq_symmpdf}. The difference lies in the scalar random variable $\hat{\mathcal{R}}$ that does no longer require to be non-negative and can be even negative; $\hat{\mathcal{R}}$ and ${\bm{\mathcal{U}}}$ are also dependent, which skews the distribution. This generalisation includes the skewed elliptical distribution in \eqref{eq_skewsto} as a special case, and is closed under affine transformation, marginalisations and even conditioning \cite{frahm2004generalized}. 

The skewness of \eqref{eq_genlpdf} arises from the dependency between $\hat{\mathcal{R}}$ and ${\bm{\mathcal{U}}}$. In other words, although ${\bm{\mathcal{U}}}$ is still distributed on a unit sphere surface, its dependency with $\hat{\mathcal{R}}$ enables varying importance along directions. Given the fact that there are only a few distributions of a scalar random variable conditioned on a vector, this flexible dependency is usually abstract and under-determined in practical modelling. More importantly, despite generalising and skewing the elliptical distribution,  this dependency also impedes the analysis of the generalised elliptical distribution, because the roles of $\hat{\mathcal{R}}$, $\bm{\Lambda}$ and ${\bm{\mathcal{U}}}$ become unclear and sometimes interchangeable. For example, in the elliptical distribution, the dispersion is uniquely dominated by $\mathbf{\Sigma}$,  while this no longer holds in  generalised elliptical distributions \cite{frahm2004generalized}, and could lead to multiple minima/maxima when modelling data in practice.

\section{Generalisation via von Mises-Fisher distribution}
\subsection{The vMF elliptical distribution}
Being distributed on a unit sphere surface, the vMF is a popular choice in directional statistics \cite{lopez2011mises, costa2014estimating, kurz2016unscented, sra2018directional}. The vMF distribution is determined by two parameters: $\bm{\mu}_v$ for the main direction and $\tau$ for the concentration (denoted as $vMF(\bm{\mu}_v,\tau)$). Therefore, it is natural and beneficial to replace $\bm{\mathcal{U}}$ in \eqref{eq_symmpdf} by the vMF distribution as a way of explicitly expressing the direction information. We thus propose a new type of generalisation on the elliptical distributions in the form
\begin{equation}\label{eq_vmfpdf}
\bm{\mathcal{X}} =^d \bm{\mu} + {\mathcal{R}}\bm{\Lambda}{\bm{\mathcal{V}}},
\end{equation}
where ${\bm{\mathcal{V}}}$ denotes a random variable satisfying the vMF distribution $vMF(\bm{\mu}_v,\tau)$. In our definition, $\mathcal{R}$ is the same as that in \eqref{eq_symmpdf}, i.e., non-negative and independent of ${\bm{\mathcal{V}}}$. More importantly, when $\tau\rightarrow 0$, the vMF distribution approaches the uniform distribution on a unit sphere $\bm{\mathcal{U}}$, and consequently \eqref{eq_vmfpdf} degenerates into the symmetric elliptical distribution. This generalisation maximally preserves the formats and desirable properties of the symmetric elliptical distribution, such as the independence and clear physical meaning of each part. In other words, in our vMF elliptical distribution, $\bm{\mu}$ closely relates to the data location, $\bm{\Lambda}$ controls the dispersion, $\mathcal{R}$ governs the tails and $\bm{\mathcal{V}}$ the directions (skewness). As shall be shown shortly, this is beneficial in both theoretical analysis and practical estimator settings.

The pdf of $\bm{\mathcal{X}}$ in \eqref{eq_vmfpdf} can be obtained in a closed-form as
\begin{equation}\label{eq_vMF_generalised}
p_{\bm{\mathcal{X}}}(\mathbf{x}) = \mathrm{det}(\mathbf{\Sigma})^{-1/2}\cdot p_{\bm{\mathcal{V}}}(\frac{\mathbf{\Sigma}^{\nicefrac{-1}{2}}(\mathbf{x}-\bm{\mu})}{\sqrt{t}}) \cdot g(t),
\end{equation}
where $t$ represents the Mahalanobis distance i.e., $t=(\mathbf{x} - \bm{\mu})^T \mathbf{\Sigma}^{-1} (\mathbf{x} - \bm{\mu})$ and $p_{\bm{\mathcal{V}}}(\cdot)$ is the pdf of vMF distribution $vMF(\bm{\mu}_v,\tau)$. We provide the proof of \eqref{eq_vMF_generalised} in Appendix-\ref{proof_eq_vMF_generalised}. 
An intuitive way of understanding our generalisation is through the fact that $vMF(\bm{\mu}_v, \tau)$ resembles a Gaussian distribution $\mathcal{N}(\bm{\mu}_v, \nicefrac{1}{\tau}\mathbf{I})$ constrained on a unit circle (especially for adequately large $\tau$). Furthermore, the Gaussian distribution $\mathcal{N}(\bm{\mu}_v, \nicefrac{1}{\tau}\mathbf{I})$, as a special case of the elliptical distribution, has the stochastic representation  $\bm{\mathcal{V}}\!\approx^d\!\bm{\mu}_v\!+\!\frac{1}{\sqrt{\tau}}\sqrt{\chi^2_M}{\bm{\mathcal{U}}}.$ Therefore, we can approximate \eqref{eq_vmfpdf} as
\begin{equation}\label{eq_vMF_stochastic}
\begin{aligned}
\bm{\mathcal{X}} &=^d \bm{\mu} + \mathcal{R}{\bm{\Lambda}}{\bm{\mathcal{V}}} \approx^d  \bm{\mu} + \mathcal{R}{\bm{\Lambda}}(\bm{\mu}_v+\frac{1}{\sqrt{\tau}}\sqrt{\chi^2_M}{\bm{\mathcal{U}}})\\
&\approx^d \bm{\mu} + {\bm{\Lambda}}\bm{\mu}_v\mathcal{R} + \frac{1}{\sqrt{\tau}}\bm{\Lambda}(\mathcal{R}\sqrt{\chi^2_M})\bm{\mathcal{U}}.
\end{aligned}
\end{equation}
Given that random variables $\{\mathcal{R}, \sqrt{\chi^2_M}, ~\bm{\mathcal{U}}\}$ are mutually  independent, $\big(\bm{\mu} +  \frac{1}{\sqrt{\tau}}\bm{\Lambda}(\mathcal{R}\sqrt{\chi^2_M})\bm{\mathcal{U}}\big)$ constitutes a symmetric elliptical distribution whilst the term ${\bm{\Lambda}}\bm{\mu}_v\mathcal{R}$ adds to the asymmetry on the direction ${\bm{\Lambda}}\bm{\mu}_v$. 

\subsection{Properties}
We now proceed to introduce the properties of the proposed vMF elliptical distribution. We first address in Proposition \ref{prop_invar} the invariance under the quadratic terms. 
\begin{prop}\label{prop_invar}
The skewness of the distribution is invariant to the quadratic term $(\bm{\mathcal{X}} - \bm{\mu})^T \mathbf{\Sigma}^{-1} (\bm{\mathcal{X}} - \bm{\mu})$.
\proof 
From \eqref{eq_vmfpdf}, we have 
\begin{equation}
(\bm{\mathcal{X}} - \bm{\mu})^T \mathbf{\Sigma}^{-1} (\bm{\mathcal{X}} - \bm{\mu}) =^d \mathcal{R}^2\bm{\mathcal{V}}^T\bm{\mathcal{V}} =^d \mathcal{R}^2.
\end{equation}

Because $\mathcal{R}$ is independent of $\bm{\mathcal{V}}$, the quadratic term is irrelevant to the skewness part of our vMF elliptical distribution.

This completes the proof.
\end{prop}
The property in Proposition \ref{prop_invar} is the basis for satisfying the distributional invariance \cite{genton2001moments}, which provides convenience in applications such as hypothesis testing and dealing with sampling bias \cite{genton2005generalized}.

Given the fact that $\nicefrac{\mathbf{\Sigma}^{\nicefrac{-1}{2}}(\bm{\mathcal{X}}-\bm{\mu})}{(\bm{\mathcal{X}}-\bm{\mu})^T\mathbf{\Sigma}^{-1}(\bm{\mathcal{X}}-\bm{\mu})}=^d\bm{\mathcal{V}}$, from \eqref{eq_vmfpdf} we can easily obtain the following result based on the proof of the Proposition \ref{prop_invar}.
\begin{prop}\label{prop_independ}
The quadratic term $(\bm{\mathcal{X}} - \bm{\mu})^T \mathbf{\Sigma}^{-1} (\bm{\mathcal{X}} - \bm{\mu})$ is independent of the whitened random variable $\nicefrac{\mathbf{\Sigma}^{\nicefrac{-1}{2}}(\bm{\mathcal{X}}-\bm{\mu})}{(\bm{\mathcal{X}}-\bm{\mu})^T\mathbf{\Sigma}^{-1}(\bm{\mathcal{X}}-\bm{\mu})}$.
\end{prop}
This independence property plays a crucial role in the calculation of robustness and other related proofs.
Furthermore, due to the neat generalisation and clear physical meaning of our vMF elliptical distribution, we can write the moments in a closed-form, with the first two moments are addressed in Proposition \ref{prop_moments}.
\begin{prop}\label{prop_moments}
Given the generalisation of \eqref{eq_vmfpdf}, we have
\begin{equation}
\begin{aligned}
\mathbb{E}[\bm{\mathcal{X}}] &= \bm{\mu} + (\rho_m(\tau)\mathbb{E}[{\mathcal{R}}]\bm{\Lambda})\bm{\mu}_v\\
\mathrm{Var}[\bm{\mathcal{X}}] &= \mathbb{E}[{\mathcal{R}}]^2\frac{\rho_m(\tau)}{\tau}\mathbf{\Sigma} \\
&~~~- \mathbb{E}[{\mathcal{R}}]^2(1-\frac{m}{\tau}\rho_m(\tau)-\rho^2_m(\tau))\bm{\Lambda}\bm{\mu}_v\bm{\mu}_v^T\bm{\Lambda}^T
\end{aligned}
\end{equation}
where $\rho_m(\tau)\in(0,1)$ is the ratio of two modified Bessel functions of the first kind, given by 
\begin{equation}
\rho_m(\tau)=\nicefrac{I_{\nicefrac{m}{2}}(\tau)}{I_{\nicefrac{m}{2}-1}(\tau)}.
\end{equation}
\proof Given the stochastic representation of \eqref{eq_vmfpdf}, we have
\begin{equation}
\mathbb{E}[\bm{\mathcal{X}}] = \bm{\mu} + \mathbb{E}[{\mathcal{R}}\bm{\Lambda}{\bm{\mathcal{V}}}].
\end{equation}
More importantly, due to the independence between $\mathcal{R}$ and $\bm{\mathcal{V}}$, we further have 
\begin{equation}
\mathbb{E}[\bm{\mathcal{X}}] = \bm{\mu} + \mathbb{E}[{\mathcal{R}}]\bm{\Lambda}\mathbb{E}[{\bm{\mathcal{V}}}].
\end{equation}
Recall that ${\bm{\mathcal{V}}}$ is the vMF parametrised as $vMF(\bm{\mu}_v,\tau)$. Its first-order moment is calculated by $\mathbb{E}[{\bm{\mathcal{V}}}] = \rho_m(\tau)\bm{\mu}_v$ \cite{hillen2017moments}. Thus, the first-order moment of the vMF elliptical distribution can be obtained as
\begin{equation}
\mathbb{E}[\bm{\mathcal{X}}] = \bm{\mu} + \big(\rho_m(\tau)\mathbb{E}[{\mathcal{R}}]\bm{\Lambda}\big)\bm{\mu}_v.
\end{equation}

Similarly, the second-order moment of the vMF distribution is given by  $\mathbb{E}[{\bm{\mathcal{V}}}{\bm{\mathcal{V}}}^T] = \nicefrac{\rho_m(\tau)}{\tau}\mathbf{I}+(1-\nicefrac{m}{\tau}\rho_m(\tau))\bm{\mu}_v\bm{\mu}_v^T$ \cite{hillen2017moments}, so that we arrive at
\begin{equation}
\begin{aligned}
\mathrm{Var}[\bm{\mathcal{X}}] &=\mathbb{E}[\bm{\mathcal{X}}\bm{\mathcal{X}}^T] - \mathbb{E}[\bm{\mathcal{X}}]\mathbb{E}[\bm{\mathcal{X}}]^T \\
&= \mathbb{E}[\bm{\mu}\bm{\mu}^T \!+\! \bm{\mu}\mathcal{R}\bm{\mathcal{V}}^T\bm{\Lambda}^T \!+\! \mathcal{R}\bm{\Lambda}\bm{\mathcal{V}}\bm{\mu}^T \!+\! \mathcal{R}^2\bm{\Lambda}\bm{\mathcal{V}}\bm{\mathcal{V}}^T\bm{\Lambda}^T]\\
&~~~~- \big(\bm{\mu}\bm{\mu}^T \!+\!  \mathbb{E}[{\mathcal{R}}]\bm{\mu}\mathbb{E}[{\bm{\mathcal{V}}}]^T\bm{\Lambda}^T \!+\! \mathbb{E}[{\mathcal{R}}]\bm{\Lambda}\mathbb{E}[{\bm{\mathcal{V}}}]\bm{\mu}^T\\
&~~~~~~~~~\! +\! \mathbb{E}[{\mathcal{R}}]^2\bm{\Lambda}\mathbb{E}[{\bm{\mathcal{V}}}]\mathbb{E}[{\bm{\mathcal{V}}}]^T\bm{\Lambda}^T\big)\\
&=\mathbb{E}[{\mathcal{R}}]^2\bm{\Lambda}\big(\mathbb{E}[{\bm{\mathcal{V}}}{\bm{\mathcal{V}}}]^T - \mathbb{E}[{\bm{\mathcal{V}}}]\mathbb{E}[{\bm{\mathcal{V}}}]^T\big)\bm{\Lambda}^T\\
&=\mathbb{E}[{\mathcal{R}}]^2\frac{\rho_m(\tau)}{\tau}\mathbf{\Sigma} \\
&~~~~- \mathbb{E}[{\mathcal{R}}]^2(1-\frac{m}{\tau}\rho_m(\tau)-\rho^2_m(\tau))\bm{\Lambda}\bm{\mu}_v\bm{\mu}_v^T\bm{\Lambda}^T
\end{aligned}
\end{equation}

This completes the proof.
\end{prop}
Furthermore, when $\tau\rightarrow 0$, the following bounds hold, according to the Taylor series of the modified Bessel function of the first kind ($I_a(b) \approx \frac{1}{\Gamma(a+1)}(\frac{b}{2})^a$ for $b\rightarrow 0$):
\begin{equation}
\begin{aligned}
\lim_{\tau\rightarrow 0}\rho_m(\tau) &= \frac{I_{\nicefrac{m}{2}}(\tau)}{I_{\nicefrac{m}{2}-1}(\tau)}= \frac{\Gamma(\nicefrac{m}{2})}{2\Gamma(\nicefrac{m}{2}+1)}\tau = 0,\\
\lim_{\tau\rightarrow 0}\frac{\rho_m(\tau)}{\tau} &= \frac{\Gamma(\nicefrac{m}{2})}{2\Gamma(\nicefrac{m}{2}+1)} = \frac{\Gamma(\nicefrac{m}{2})}{2(\nicefrac{m}{2})\Gamma(\nicefrac{m}{2})} = \frac{1}{m}.\\
\end{aligned}
\end{equation}
Therefore, from Proposition \ref{prop_moments}, when $\tau\rightarrow 0$, we have $\mathbb{E}[\bm{\mathcal{X}}] = \bm{\mu}$ and $\mathrm{Var}[\bm{\mathcal{X}}] = \nicefrac{\mathbb{E}[\mathcal{R}^2]}{m}\mathbf{\Sigma}$, which are exactly the moments of the symmetric elliptical distributions \cite{frahm2004generalized}.

\subsection{Optimisation}
Given the i.i.d. nature of the samples $\{\mathbf{x}_i\}_{i=1}^n$, our vMF elliptical distribution can be estimated by maximising the log-likelihood (MLL), which is equivalent to minimising the Kullback-Leibler divergence between the parametric and empirical distributions. It also needs to be pointed out that other distance measures can be employed in estimation, such as the Wasserstein distance, a subject of future work.

For notational simplicity, we shall denote the whitened term as $\mathbf{z}_i = \frac{\mathbf{\Sigma}^{\nicefrac{-1}{2}}(\mathbf{x}_i-\bm{\mu})}{\sqrt{t}}$ and the quadratic term as  $t_i=(\mathbf{x}_i - \bm{\mu})^T \mathbf{\Sigma}^{-1} (\mathbf{x}_i - \bm{\mu})$. Then, due to the simple form of the pdf of our generalised distribution, we can omit the constant term and write the log-likelihood from \eqref{eq_vMF_generalised} as
\begin{equation}\label{eq_loglikeli1}
\begin{aligned}
L(\bm{\mu}, \mathbf{\Sigma}, & ~\bm{\mu}_v, \tau) \propto -\frac{n}{2}\ln\mathrm{det}(\mathbf{\Sigma}) + n(\frac{m}{2}-1)\ln\tau \\
&- n\ln I_{\nicefrac{m}{2}-1}(\tau) + \sum_{i=1}^n\tau\bm{\mu}_v^T\mathbf{z}_i + \sum_{i=1}^n\ln g(t_i)
\end{aligned}
\end{equation}
Furthermore, as $||\bm{\mu}_v||=1$, this leads to constrained optimisation, whereby we reparametrise \eqref{eq_loglikeli1} by $\mathbf{v}=\tau\bm{\mu}_v$ (where $\tau = ||\mathbf{v}||_2$ and $\bm{\mu}_v=\nicefrac{\mathbf{v}}{||\mathbf{v}||_2}$) as
\begin{equation}
\begin{aligned}
L(\bm{\mu}, \mathbf{\Sigma}, & ~\mathbf{v}) \propto -\frac{n}{2}\ln\mathrm{det}(\mathbf{\Sigma}) + n(\frac{m}{2}-1)\ln||\mathbf{v}||_2 \\
&- n\ln I_{\nicefrac{m}{2}-1}(||\mathbf{v}||_2) + \sum_{i=1}^n\mathbf{v}^T\mathbf{z}_i + \sum_{i=1}^n\ln g(t_i).
\end{aligned}
\end{equation}

By taking the derivatives with respect to $\bm{\mu}$, $\mathbf{\Sigma}$ and $\mathbf{v}$, and setting them to $\mathbf{0}$, we obtain the following optimum, with the detailed proof given in Appendix \ref{proof_eq_optimum}.
\begin{equation}\label{eq_optimum}
\begin{aligned}
\bm{\mu}^* &= \frac{\sum_{i=1}^n\big((\frac{\mathbf{v}^T\mathbf{z}_i}{t_i} - 2\psi(t_i))\mathbf{x}_i - \frac{1}{\sqrt{t_i}}\mathbf{\Sigma}^{\frac{1}{2}}\mathbf{v}\big)}{\sum_{i=1}^n(\frac{\mathbf{v}^T\mathbf{z}_i}{t_i} - 2\psi(t_i))} \\
\mathbf{\Sigma}^* &= \frac{\sum_{i=1}^n\big((\frac{\mathbf{v}^T\mathbf{z}_i}{t_i}-2\psi(t_i))(\mathbf{x}_i-\bm{\mu})(\mathbf{x}_i-\bm{\mu})^T - \mathbf{\Sigma}\mathbf{z}_i\mathbf{v}^T\big)}{n} \\
\mathbf{v}^* &= \rho_m^{-1}(||\frac{1}{n}\sum_{i=1}^n\mathbf{z}_i||_2)\frac{\sum_{i=1}^n\mathbf{z}_i}{||\sum_{i=1}^n\mathbf{z}_i||_2}
\end{aligned}
\end{equation}
In \eqref{eq_optimum}, $\rho_m^{-1}(\cdot)$ is an inverse function of $\rho_m(\cdot)$ and $\psi(t) = \frac{\nicefrac{\partial g(t)}{\partial t}}{g(t)}$. Moreover, when $\tau\rightarrow 0$, $\mathbf{v}^*\rightarrow\mathbf{0}$, and the optimal $\bm{\mu}$ and $\mathbf{\Sigma}$ also represent the optimum of the symmetric elliptical distribution.

\section{Numerical results}
\begin{figure}[!htb]
	\centering
	\subfigure[vMF Gaussian]{\includegraphics[width=0.4\textwidth]{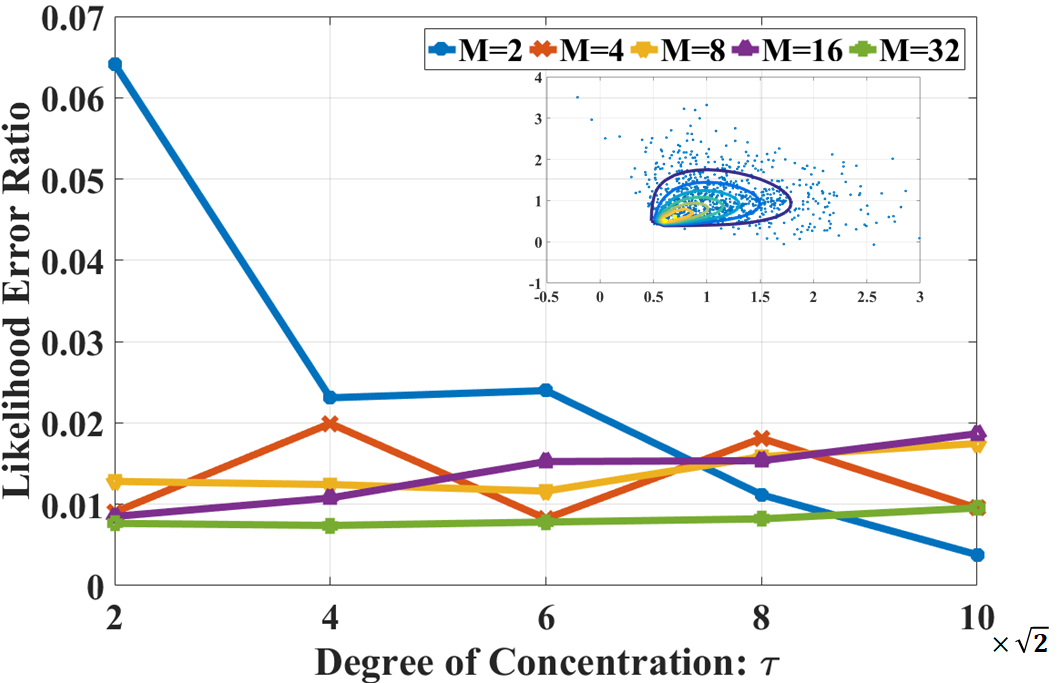}}
	\subfigure[vMF Cauchy]{\includegraphics[width=0.4\textwidth]{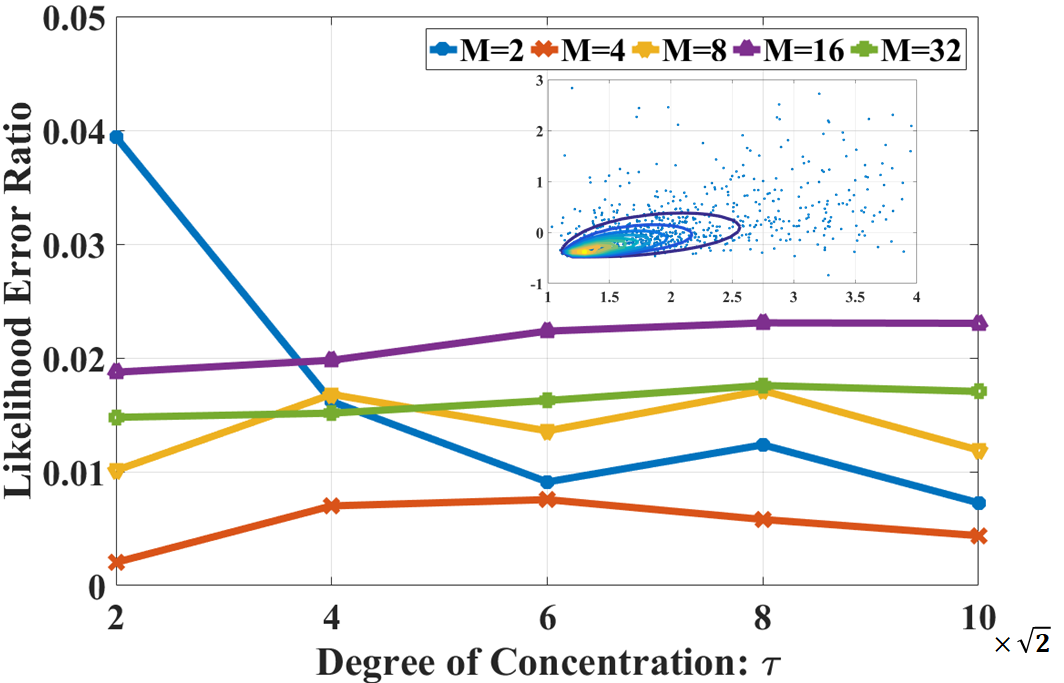}}
	\vspace{-.5em}
	\caption{Log-likelihood error ratios with respect to different dimensions, $m$, and degrees of concentration, $\tau$, in estimating the proposed: (a) vMF Gaussian and (b) vMF Cauchy distributions. One example per distribution of randomly generated samples from the stochastic representations of \eqref{eq_vmfpdf} is also provided in the upper right panel. For the convenience of illustration, we show the main part of vMF Cauchy samples whilst ignoring some highly scattered samples.}\label{figure_aucap}
	\vspace{-.5em}
\end{figure}

This section supports the analysis through experimental assessment of properties of the proposed vMF elliptical distribution, together with its estimation method. In our experiments, we tested 5 degrees of concentration, i.e., $\tau=\{2\sqrt{2}, 4\sqrt{2}, 6\sqrt{2}, 8\sqrt{2}, 10\sqrt{2}\}$, together with $5$ different dimensions ($m=\{2,4,8,16,32\}$), under two types of distributions, i.e., vMF Gaussian and vMF Cauchy distributions, thus having $5\times5\times2=50$ test cases. For each test case, we generated $1,000$ synthetic samples from distributions of randomly chosen parameters ($\bm{\mu}, \bm{\mu}_v, \mathbf{\Sigma}$). The only controlled parameter was the eccentricity of the distribution, which is the ratio of the maximal eigenvalue over the minimal eigenvalue of $\mathbf{\Sigma}$; we set it smaller than $4$ to avoid the majority of samples being aligned on low-dimensional spaces. For the synthetic data, in each test case, we performed estimation over $10$ trails, with random initialisations. Standard steepest gradient descent was used to update $\bm{\mu}$ and $\mathbf{\Sigma}$, with the learning rate of $0.01$, while $\bm{\mu}_v$ and $\tau$ were optimised by a truncated Newton method proposed in \cite{sra2018directional}. The mean values of likelihood error ratios are plotted in Figure \ref{figure_aucap}, with the likelihood error ratio  defined as $r = \nicefrac{|l_{est}-l_{true}|}{l_{true}},$
while $l_{est}$ denotes the log-likelihood given by the estimated parameters and $l_{true}$ for the true log-likelihood. 

From Figure \ref{figure_aucap}, we observe that different from other generalised elliptical distributions, generating our vMF elliptical distributions is straightforward to implement and easy to control. More importantly, by virtue of its simple estimation method, the error ratios of estimating both vMF Gaussian and Cauchy distributions were both lower than $0.07$, and even lower than $0.03$ except for the case $m=2$. However, the obtained small error ratios also verify that the proposed vMF elliptical distribution is not only easy to generate but is also stable. Moreover, Figure \ref{figure_aucap} also shows that increasing either the dimensions $m$ or the degrees of concentration $\tau$ did not enlarge the estimated error, thus validating the consistency of our simple yet robust and effective estimation. Employing advanced numerical solvers such as conjugate gradient descent and trust region method could further improve the estimation, a subject of future work.

\section{Conclusion}
We have proposed a novel generalised elliptical distribution, termed vMF elliptical distribution, in order to generalise symmetric elliptical distributions and equip them with a physical meaning in the analysis, simple representations and the ability to represent general skewed distributions. This has been achieved in an intuitive manner, starting from the stochastic representation, which has enabled the proposed vMF elliptical distribution to exhibit many desirable properties, together with an explicit pdf. We have also introduced optimal parameter estimation into the vMF elliptical distributions, and have experimentally validated that even a basic steepest gradient descent method can achieve consistent and stable low estimation errors. This opens new avenues for both future applications and research on skewed distributions, especially in the context of probabilistic learning machines.

\section{Appendix}
\subsection{Proof of \eqref{eq_vMF_generalised}}\label{proof_eq_vMF_generalised}
Notice that $\mathcal{R}$ is independent of $\bm{\mathcal{V}}$, so that the following holds
\begin{equation}
p_{\{\mathcal{R}, \bm{\mathcal{V}}\}}(r,\mathbf{v}) = p_{\mathcal{R}}(r)\cdot p_{\bm{\mathcal{V}}}(\mathbf{v})
\end{equation}

We next define a random variable by a multiplication $\bm{\mathcal{Y}} = f(\mathcal{R}, \bm{\mathcal{V}}) = \mathcal{R}\cdot\bm{\mathcal{V}}$. It needs to be pointed out that $\bm{\mathcal{V}}$ is a unit random vector, which means that the multiplication function $f$ is an injection due to $\{\mathcal{R}, \bm{\mathcal{V}}\} = f^{-1}(\bm{\mathcal{Y}}) = \{||\bm{\mathcal{Y}}||_2, \nicefrac{\bm{\mathcal{Y}}}{||\bm{\mathcal{Y}}||_2}\}$. 
%The Jacobian of this transform is given by
%\begin{equation}
%J_{(m+1)\times m} = \begin{bmatrix}
%\bm{\mathcal{U}}_v \\
%\mathcal{R}\mathbf{I}_m
%\end{bmatrix}
%\end{equation}
Given the change of variables formula \cite{ben1999change}, we have that the matrix volume is $||\mathbf{y}||_2^{-(m-1)}$. Thus, the pdf of $\mathcal{Y}$ can be obtained as 
\begin{equation}
\begin{aligned}
p_{\bm{\mathcal{Y}}}(\mathbf{y}) &= p_{\{\mathcal{R},  \bm{\mathcal{V}}\}}(||\mathbf{y}||_2, \frac{\mathbf{y}}{||\mathbf{y}||_2})\cdot||\mathbf{y}||_2^{-(m-1)} \\
&= p_{\mathcal{R}}(||\mathbf{y}||_2)\cdot p_{\bm{\mathcal{V}}}(\frac{\mathbf{y}}{||\mathbf{y}||_2})\cdot||\mathbf{y}||_2^{-(m-1)}
\end{aligned}
\end{equation}
By employing the linear transform, $\bm{\mathcal{X}} = \bm{\mu}+\mathbf{\Sigma}^{\nicefrac{1}{2}}\bm{\mathcal{Y}}$, the pdf of our vMF elliptical distribution can be obtained by substituting $t = (\mathbf{x}-\bm{\mu})^T\mathbf{\Sigma}^{-1}(\mathbf{x}-\bm{\mu}) = ||\mathbf{y}||_2^2$, as
\begin{equation}
p_{\bm{\mathcal{X}}}(\mathbf{x}) = \mathrm{det}(\mathbf{\Sigma})^{-\frac{1}{2}}\cdot p_{\bm{\mathcal{V}}}(\frac{\mathbf{\Sigma}^{\nicefrac{-1}{2}}(\mathbf{x}-\bm{\mu})}{\sqrt{t}}) \cdot \underbrace{t^{-\frac{m-1}{2}}p_\mathcal{R}\big(\sqrt{t}\big)}_{g(t)}
\end{equation}

\subsection{Proof of \eqref{eq_optimum}}\label{proof_eq_optimum}
%We first calculate the optimum of $\mathbf{v}$ as follows,
%\begin{equation}
%\begin{aligned}
%\frac{1}{n}\frac{\partial L(\bm{\mu}, \mathbf{\Sigma}, \mathbf{v})}{\partial \mathbf{v}} &= (\frac{m}{2}-1)\frac{\mathbf{v}}{||\mathbf{v}||^2_2} + \frac{1}{n}\sum_{i=1}^n\mathbf{z}_i\\
%&- \frac{1}{I_{\nicefrac{m}{2}-1}(||\mathbf{v}||_2)}\frac{\partial I_{\nicefrac{m}{2}-1}(||\mathbf{v}||_2)}{\partial \mathbf{v}}.
%\end{aligned}
%\end{equation}
By using the relationship of the modified Bessel function of the first kind $\nicefrac{\partial I_a(b)}{\partial b} = I_{a+1}(b) + \nicefrac{a}{b}I_{a}(b)$, we can simplify the derivative of $\mathbf{v}$ as 
\begin{equation}\label{eq_proof_eq_optimum1}
\begin{aligned}
&\frac{1}{n}\frac{\partial L(\bm{\mu}, \mathbf{\Sigma}, \mathbf{v})}{\partial \mathbf{v}} = (\frac{m}{2}-1)\frac{\mathbf{v}}{||\mathbf{v}||^2_2} + \frac{1}{n}\sum_{i=1}^n\mathbf{z}_i\\
&- \frac{1}{I_{\nicefrac{m}{2}-1}(||\mathbf{v}||_2)}\frac{\mathbf{v}}{||\mathbf{v}||_2}\big(I_{\nicefrac{m}{2}}(||\mathbf{v}||_2)+\frac{\frac{m}{2}-1}{||\mathbf{v}||_2}I_{\nicefrac{m}{2}-1}(||\mathbf{v}||_2)\big)\\
&~~~~~~~~~~~~~~~~~= -\frac{\mathbf{v}}{||\mathbf{v}||_2}\frac{I_{\nicefrac{m}{2}}(||\mathbf{v}||_2)}{I_{\nicefrac{m}{2}-1}(||\mathbf{v}||_2)} + \frac{1}{n}\sum_{i=1}^n\mathbf{z}_i
\end{aligned}
\end{equation}
By setting the derivative in \eqref{eq_proof_eq_optimum1} to $\mathbf{0}$, we arrive at the optimal $\mathbf{v}$ in the form
\begin{equation}
\mathbf{v}^* = \frac{||\mathbf{v}||_2}{\rho_m(||\mathbf{v}||_2)}\frac{1}{n}\sum_{i=1}^n\mathbf{z}_i = \rho_m^{-1}(||\frac{1}{n}\sum_{i=1}^n\mathbf{z}_i||_2)\frac{\sum_{i=1}^n\mathbf{z}_i}{||\sum_{i=1}^n\mathbf{z}_i||_2}
\end{equation}
The optimal $\bm{\mu}$ can be calculated in a similar way
\begin{equation}
\begin{aligned}
0&=\frac{\partial L(\bm{\mu}, \mathbf{\Sigma}, \mathbf{v})}{\partial \bm{\mu}}=\sum_{i=1}^n\big(\frac{-\mathbf{\Sigma}^{-\frac{1}{2}}\mathbf{v}}{t_i^{0.5}}+\frac{\mathbf{v}^T\mathbf{z}_i\mathbf{\Sigma}^{-1}(\mathbf{x}_i-\bm{\mathbf{\mu}})}{t_i}\big) \\
& - 2\psi(t_i)\mathbf{\Sigma}^{-1}(\mathbf{x}_i-\bm{\mathbf{\mu}}),
\end{aligned}
\end{equation}
with the optimum at
\begin{equation}
\bm{\mu}^* = \frac{\sum_{i=1}^n\big((\frac{\mathbf{v}^T\mathbf{z}_i}{t_i} - 2\psi(t_i))\mathbf{x}_i - \frac{1}{\sqrt{t_i}}\mathbf{\Sigma}^{\frac{1}{2}}\mathbf{v}\big)}{\sum_{i=1}^n(\frac{\mathbf{v}^T\mathbf{z}_i}{t_i} - 2\psi(t_i))}
\end{equation}

To find the optimum of $\mathbf{\Sigma}$, it is convenient to decompose $\mathbf{\Sigma}$ into $\bm{\Lambda}\bm{\Lambda}^T$ by the Cholesky decomposition because the square root of the positive definite matrix $\mathbf{\Sigma}$ is guaranteed. Then, by taking the derivatives with regard to $\bm{\Lambda}$, we have
\begin{equation}
\begin{aligned}
0&=\frac{\partial L(\bm{\mu}, \mathbf{\Sigma}\!=\!\bm{\Lambda}\bm{\Lambda}^T, \mathbf{v})}{\partial \bm{\Lambda}} = -n\bm{\Lambda}^{-T} \\
&+ \sum_{i=1}^n\big(\frac{1}{t_i}\mathbf{v}^T\mathbf{z}_i \mathbf{\Sigma}^{-1}(\mathbf{x}_i-\bm{\mu})(\mathbf{x}_i-\bm{\mu})^T\bm{\Lambda}^{-T} - \mathbf{z}_i\mathbf{v}^T\bm{\Lambda}^{-T}\big)\\
& - \sum_{i=1}^n\big(2\psi(t_i)\mathbf{\Sigma}^{-1}(\mathbf{x}_i-\bm{\mu})(\mathbf{x}_i-\bm{\mu})^T\bm{\Lambda}^{-T}\big)
\end{aligned}
\end{equation}
After some simplifications, the optimal $\mathbf{\Sigma}$ can be easily obtained in a closed-form as
\begin{equation}
\mathbf{\Sigma}^* = \frac{\sum_{i=1}^n\big((\frac{\mathbf{v}^T\mathbf{z}_i}{t_i}-2\psi(t_i))(\mathbf{x}_i-\bm{\mu})(\mathbf{x}_i-\bm{\mu})^T - \mathbf{\Sigma}\mathbf{z}_i\mathbf{v}^T\big)}{n}
\end{equation}

\bibliographystyle{IEEEtran}
\bibliography{IEEEfull,ShengxiLi}
\end{document}